%% file: main.tex
\documentclass[letterpaper,twocolumn,anonymous,10pt]{article}
\usepackage{usenix2019_v3}
\usepackage{tikz}
\usepackage{amsmath}

\usepackage{filecontents}
\usepackage{xspace}
\usepackage{threeparttable}
\usepackage{makecell}
\usepackage{url}
\usepackage{cleveref}

\usepackage{tikz}
\usepackage{threeparttable}
\usepackage{soul}
\usepackage{colortbl}
\usepackage{multirow}
\usepackage{tabularx}
\usepackage{amsmath,pifont, mathtools}
\usepackage{amsmath}
\usepackage[overload]{empheq}
\usepackage{nccmath}
\usepackage{cases}
\usepackage{bbding}
\usepackage{graphicx}
\usepackage{pifont}
\usepackage{wasysym}
\usepackage{mathtools}
\usepackage{float}
\usepackage{subcaption}
\usepackage{adjustbox}
\usepackage{makecell}%
\usepackage{tikz}
\usepackage{booktabs}
\usepackage{cleveref}
\usepackage{comment}
\usepackage[normalem]{ulem}
\usepackage{enumitem}
\usepackage{listings}
\usepackage{color}
\usepackage{textcomp}
\usepackage{balance}

\usepackage{booktabs}
\usepackage{caption}
\usepackage{tikz}
\usepackage{xspace}
\usepackage{pifont}
\newcommand{\xmark}{\ding{55}}
\usepackage{url}

\usepackage[font=small]{caption}
\definecolor{red}{rgb}{1,0,0}
\definecolor{gray}{rgb}{0.5,0.5,0.5}

\crefformat{subsection}{§#1}
\crefformat{section}{§#1}

\usepackage{bbding}
\usepackage{amssymb}

\newcommand\hmm[1]{\ifnum\ifhmode\spacefactor\else2000\fi>1000 \uppercase{#1}\else#1\fi}
\newcommand{\sys}{Frontier\xspace}

\newcommand{\at}{\texttt{Attention}\xspace}
\newcommand{\Gg}{\texttt{GroupedGEMM}\xspace}

\usepackage{enumitem} 
\setlist{noitemsep,topsep=2pt,parsep=2pt,partopsep=0pt}
\usepackage{cleveref}
\crefname{section}{}{\S\S}
\crefdefaultlabelformat{\S#2#1#3}
\begin{document}


\title{\sys: Simulating the Next Generation of LLM Inference Systems}

\author{
{\rm Yicheng Feng$^{\text{1}}$}\enskip
{\rm Xin Tan$^{\text{1}}$}\enskip 
{\rm Kin Hang Sew$^{\text{1}}$}\enskip
{\rm Yimin Jiang$^{\text{2}}$}\enskip
{\rm Yibo Zhu$^{\text{2}}$}\enskip
{\rm Hong Xu$^{\text{1}}$}\enskip
\\
\\
$^{\text{1}}$The Chinese University of Hong Kong\enskip\enskip
$^{\text{2}}$StepFun\enskip
}

\maketitle

\input{abstract}
\input{introduction}
\input{motivation}

\input{design}
\input{evaluation}

\input{conclusion}

\clearpage
\bibliographystyle{plain}
\bibliography{reference}


\end{document}

%% file: abstract.tex

\begin{abstract}

Large Language Model (LLM) inference is growing increasingly complex with the rise of Mixture-of-Experts (MoE) models and disaggregated architectures that decouple components like prefill/decode (PD) or attention/FFN (AF) for heterogeneous scaling.
Existing simulators, however, fall short in modeling the system-level complexities of distributed serving, and thus are unable to capture the intricate system dynamics of these emerging paradigms.
We present \sys, a high-fidelity simulator designed from the ground up for this new landscape. \sys introduces a unified framework to model both co-located and disaggregated systems, providing native support for MoE inference with expert parallelism (EP). It enables the simulation of complex workflows like cross-cluster expert routing and advanced pipelining strategies for latency hiding. To ensure fidelity and usability, \sys incorporates refined operator models for improved accuracy. \sys empowers the community to design and optimize the future of LLM inference at scale. 

\end{abstract}

%% file: introduction.tex

\section{Introduction} 
\label{sec:introduction}

The demand for scalable and cost-effective Large Language Model (LLM) serving is driving a fundamental shift away from traditional, co-located deployments. The industry is actively exploring next-generation paradigms to improve performance and efficiency~\cite{mitra2025beyond,stepfun2025step3largeaffordablemodelsystem,zhu2025megascale,zhong2024distserve,liu2024deepseek,guo2025deepseek,agrawal2024taming,Dynamo}. These include \textit{Mixture-of-Experts (MoE) models}~\cite{li2023accelerating,mitra2025beyond,singh2023hybrid,liu2024deepseek}, which employ sparse activation to scale parameter counts sub-linearly with compute costs, and \textit{disaggregated architectures}. Disaggregation splits the inference process into distinct computational stages such as separating compute-intensive prefill from memory-bound decode~\cite{zhong2024distserve,Dynamo} (PD disaggregation) or decoupling Attention from FFN computations~\cite{stepfun2025step3largeaffordablemodelsystem,zhu2025megascale} (AF disaggregation)—to optimize resource usage for each. Such approaches offer a promising path to improving the critical trade-off between system throughput and user-perceived interactivity, often referred to as the Pareto frontier~\cite{mitra2025beyond,agrawal2024vidur, zhu2025megascale}.

While promising, these advanced architectures introduce unprecedented complexity. Designing a disaggregated system, for example, requires navigating a vast and intricate search space of model partitioning, concurrency control, and dynamic rate matching between specialized hardware pools. Similarly, MoE models introduce systemic challenges of token load imbalance across experts and expensive collective communication for result aggregation~\cite{liu2024deepseek}. Optimizing these systems through empirical, trial-and-error experimentation on real hardware is prohibitively expensive and time-consuming, given the immense configuration space~\cite{agrawal2024vidur,mitra2025beyond,cho2024llmservingsim}. For example, identifying the optimal serving configuration for a standard 72B dense model in a 16-GPU co-located cluster setting can consume around 18,000 GPU-hours—amounting to a cost of over \$93,000~\cite{agrawal2024vidur}.

High-fidelity simulation is a promising tool for tackling this complexity~\cite{agrawal2024vidur,cho2024llmservingsim}. However, state-of-the-art simulators like Vidur~\cite{agrawal2024vidur} are built around a replica-centric abstraction, which is fundamentally misaligned with the architecture of fully distributed and disaggregated LLM inference systems. This traditional design views the system as a pool of homogeneous, self-contained replicas, reducing the primary challenge to load-balancing requests among them. This core assumption is broken by disaggregated and MoE architectures, where inference is no longer a monolithic task but a multi-stage workflow orchestrated across specialized, heterogeneous clusters. The replica-centric model lacks the native primitives to represent this workflow, including inter-cluster routing, data transfer (e.g., KV-Cache), and complex synchronization. The critical abstraction has thus shifted from managing a pool of replicas to orchestrating the flow of a request through a distributed system—a concept prior simulators cannot natively represent.

Accurately simulating these new paradigms requires addressing three fundamental challenges.

\textit{First, the challenge of intra-node modeling fidelity: pushing the accuracy and completeness of operator-level modeling.}
The foundation of any simulator is its ability to accurately model the execution within a single computational graph, a domain where existing simulators have primarily focused~\cite{agrawal2024vidur,cho2024llmservingsim}. However, this foundation is cracking under the pressure of modern workloads and architectures. Two critical gaps have emerged: (1) Inaccurate modeling for modern workloads: The predictive power of existing operator models, particularly for \at, degrades significantly on batches with high variance in sequence lengths. For instance, because Vidur's attention model oversimplifies the operator's runtime characteristics, we found that it can exhibit an error of over 55\% (0.151ms vs. 0.340ms) for a single FlashAttention operation on a batch of 72 requests with skewed lengths. Such shortcomings in capturing operator workload dynamics can severely undermine simulation accuracy—particularly in multi-batch scenarios.
(2) Incomplete modeling for new paradigms: Crucial computational patterns from emerging architectures, such as the heterogeneous \Gg in MoE models, are simply not accounted for, leaving a significant blind spot in performance prediction.

\textit{Second, the challenge of inter-node orchestration: creating a new simulation abstraction for distributed, multi-stage workflows.}
Even with perfect intra-node models, fully distributed systems fundamentally break the traditional replica-centric abstraction, recasting inference as a distributed workflow across specialized, independent clusters—a "system-of-systems". This demands a new simulation paradigm capable of modeling inter-node coordination. For instance, simulating PD disaggregation requires capturing the producer-consumer dynamics, where the prefill stage's output rate is constrained by the decode stage's memory availability via system-level backpressure~\cite{mitra2025beyond,zhong2024distserve,Dynamo}. Likewise, simulating AF disaggregation necessitates modeling a tightly-coupled, fine-grained pipeline, where the end-to-end latency is determined by the critical path of an event dependency graph that spans multiple clusters~\cite{mitra2025beyond,stepfun2025step3largeaffordablemodelsystem,zhu2025megascale}. Existing simulators lack the native primitives to express these stateful, inter-dependent workflows, creating a critical gap in our ability to reason about the performance of disaggregated architectures.

\textit{Third, the challenge of system-level practical constraints: modeling the diverse and dynamic policies of real-world inference engines.}
A physically deployed system's performance is ultimately governed by the software policies of its serving engine. Different engines (e.g., vLLM~\cite{kwon2023efficient}, SGLang~\cite{zheng2024sglang}, TensorRT-LLM~\cite{TensorRT}) implement a wide array of strategies for dynamic batching, request scheduling, and memory management (e.g., PagedAttention~\cite{kwon2023efficient}). These policies create complex, dynamic behaviors that significantly impact performance but are often abstracted away in current simulators. A truly practical simulation framework must treat these system-level policies as first-class citizens, allowing researchers to plug in, compose, and evaluate different strategies—from a specific batching algorithm to a novel memory management scheme. 

To address these fundamental simulation challenges, we present \textit{\sys}, \textbf{the first simulation framework designed to systematically explore the design space of next-generation LLM inference systems, i.e., systems featuring disaggregated and MoE architectures}. Our primary contribution is a novel stage-centric simulation architecture that fundamentally departs from the traditional replica-based abstraction. This new abstraction provides the native primitives required to model complex, distributed workflows, enabling \sys to capture critical inter-node dynamics.
\sys is grounded in a high-fidelity execution predictor capable of modeling data-dependent micro-workflows like MoE straggler effects, and is exposed through a modular framework with pluggable modules for exploring diverse, system-level policies from real-world inference engines. 
We plan to open source \sys to the community.







%% file: motivation.tex

\begin{table}[t]
\centering
\small

\setlength{\tabcolsep}{5pt}  
\renewcommand{\arraystretch}{1.}
\begin{tabular}{@{}l|>{\centering\arraybackslash}p{0.65cm}>{\centering\arraybackslash}p{0.65cm}|ccc|c@{}}
\toprule
\small
\multirow{3}{*}{Simulator} &
\multicolumn{2}{c|}{Disagg.} &
\multicolumn{3}{c|}{Parallelism} &
\multirow{3}{*}{Sched.} \\
\cmidrule(lr){2-3} \cmidrule(lr){4-6}
& \multicolumn{1}{c}{PD} & \multicolumn{1}{c|}{AF} & PP/TP & DP & EP & \\
\midrule
LLMServingSim~\cite{cho2024llmservingsim} & \xmark & \xmark & \checkmark & \xmark & \xmark & \xmark \\
Vidur~\cite{agrawal2024vidur} & \xmark & \xmark & \checkmark & \xmark & \xmark & -- \\
\textbf{\sys (ours)} & \textbf{\checkmark} & \textbf{\checkmark} & \textbf{\checkmark} & \textbf{\checkmark} & \textbf{\checkmark} & \textbf{\checkmark} \\
\bottomrule
\end{tabular}
\caption{Comparison of state-of-art inference simulators. PD: Prefill/Decode disaggregation, AF: Attention/FFN disaggregation, EP: Expert Parallelism. Sched: diverse advanced batching/memory scheduling. \checkmark: full support, \xmark: no support. \textbf{--}: partial or conditional support.}
\label{tab:work_compare}
\end{table}

\section{Background and Motivation} 
\label{sec:motivation}

\subsection{The Next-Generation Inference Paradigms} 
\label{subsec:background}
To overcome the scaling and efficiency limitations of traditional co-located deployments, the field is converging on two primary architectural paradigms that restructure the inference process.

\noindent\textbf{MoE architecture.}
MoE models replace dense FFN layers with numerous "expert" FFNs, only a subset of which are activated for each token by a router network~\cite{li2023accelerating,mitra2025beyond,singh2023hybrid,liu2024deepseek}. This design allows for a massive increase in parameter count with only a sub-linear rise in computational cost. 

\vspace{0.1em}
\noindent\textbf{Inference disaggregation.} 
This paradigm exploits the distinct computational profiles of different inference phases by assigning them to specialized, independent hardware clusters. This is applied at a coarse grain by separating compute-bound prefill from memory-bandwidth-bound decode (PD disaggregation)~\cite{zhong2024distserve,Dynamo}, or at a finer grain by decoupling the attention and FFN computations (AF disaggregation)~\cite{stepfun2025step3largeaffordablemodelsystem,zhu2025megascale}. 

\subsection{The Simulation Gap of Existing Simulators} 
\label{subsec:gap}
Existing LLM inference simulators~\cite{agrawal2024vidur,cho2024llmservingsim} exhibit a substantial simulation gap when applied to emerging paradigms. This gap aligns with the challenges discussed in ~\cref{sec:introduction} and is evident in three key dimensions. Table~\ref{tab:work_compare} contrasts \sys with prior simulators. Several intra-framework simulators~\cite{mitra2025beyond,zhong2024distserve}, likely based on simplified roofline models, suffer from low fidelity.


%% file: design.tex

\section{\sys Design}
\label{sec:design}
\sys models the disaggregated and MoE inference based on the key designs and insights of mainstream frameworks (e.g., vLLM~\cite{kwon2023efficient}, SGLang~\cite{zheng2024sglang}, TensorRT-LLM~\cite{TensorRT}) and works (e.g., DistServe~\cite{zhong2024distserve}, Step-3~\cite{stepfun2025step3largeaffordablemodelsystem}, MegaScale-Infer~\cite{zhu2025megascale}).
\sys adheres to the event-driven and modular design principles established by Vidur~\cite{agrawal2024vidur}.

\subsection{Architecture Overview}
\label{subsec:overview}

\begin{figure}[t] 
    \centering 
    \includegraphics[width=0.95\linewidth]{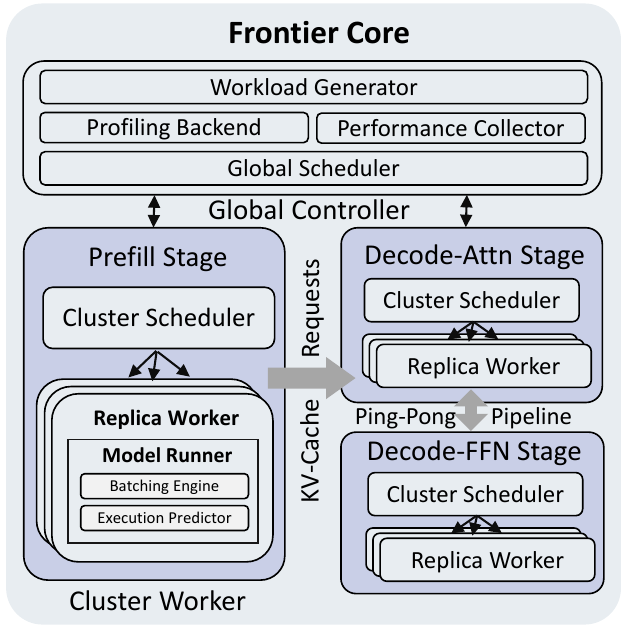} 
    \vspace{-1mm}
    \caption{
    \sys architecture overview. 
    }
    \label{fig:archi}
\end{figure}

The \textbf{\sys Core}, depicted in Figure~\ref{fig:archi}, is a hierarchical system designed to model the complex interactions within and between specialized compute clusters. The design comprises a central orchestration entity, the \texttt{GlobalController}, and a collection of modular \texttt{ClusterWorker}s, providing a principled framework for simulating the system-of-systems nature of modern LLM serving.

\vspace{0.1em}
\noindent\textbf{Global Controller.} The \texttt{GlobalController} is the stateful orchestrator of inter-stage workflows, essential for modeling disaggregated systems. It manages the end-to-end lifecycle of requests by coordinating events between independent \texttt{ClusterWorker}s, supported by integrated modules for workload generation and performance data. Its key role is managing complex, state-dependent interactions: in PD disaggregation, it models system-level backpressure by initiating KV-Cache transfers only upon receiving memory availability signals; in AF disaggregation, it orchestrates the event dependency graph for the fine-grained pipeline.

\vspace{0.1em}

\noindent\textbf{Cluster Worker.} A \texttt{ClusterWorker} is the fundamental abstraction for a specialized hardware cluster (e.g., a prefill or attention cluster), containing a \texttt{ClusterScheduler} and a pool of \texttt{ReplicaWorker}s. The \texttt{ClusterScheduler} manages local resources and participates in inter-stage coordination, such as signaling memory availability for pull-based transfers in PD disaggregation or managing micro-batch handoffs in the AF pipeline.

\vspace{0.1em}

\noindent\textbf{Replica Worker.} The \texttt{ReplicaWorker} simulates a single model instance, with its core logic encapsulated in the \texttt{Execution} \texttt{Predictor}. Moving beyond monolithic operators, the predictor's key feature is its ability to decompose a logical layer into a data-dependent micro-workflow of events. This is critical for MoE simulation, where it models the gating decision to generate a token-to-expert assignment map and simulates expert computation as a set of heterogeneous tasks. By taking the maximum of these varied task times, the \texttt{ExecutionPredictor} natively captures the performance impact of token load imbalance and the resulting straggler effects.

\subsection{Accurate Operator Runtime Prediction}
\label{subsec:op_simulation}
\noindent\textbf{Challenges.} Unlike simple matrix multiplication (e.g., GEMM), where runtime mainly depends on input size, many operators have more complex input characteristics. For example, \at can involve widely varying sequence lengths within a batch, and \Gg often faces imbalanced internal workloads, making runtime prediction significantly more challenging.
Existing methods like Vidur simplify estimation by using a single proxy length (typically the square root of batch sequence lengths), but this overlooks important factors affecting actual kernel execution. In practice, kernel execution involves partitioning and tiling computations—processes that become much more complex and less efficient with input heterogeneity, leading to phenomena such as wave quantization.

\noindent\textbf{Finer-grained modeling.} To overcome these challenges, we employ fine-grained modeling tailored to the computation patterns and input characteristics of specific operators. For \at, we utilize a rich set of features—including aggregate and distributional statistics of sequence lengths—to train an ML model (e.g. random forest~\cite{breiman2001randomforest}) that more accurately captures workload dynamics, particularly under high input variance. Likewise, for \Gg in MoE, we extract features that reflect both input properties and expert load distribution, such as token counts, expert number, model dimensions, expert selection ratios, and various load balance metrics. Such a comprehensive approach yields robust and precise predictions, even for workloads with highly variable characteristics.

\vspace{-0.5em}

\vspace{-0.4em}
\subsection{Modeling Disaggregated and MoE Infer.}
\label{subsec:disaggregated_simulation}
\noindent\textbf{PD disaggregation workflow simulation.}
The fundamental problem of simulating PD disaggregation is to \textit{accurately model the producer-consumer dynamics between two specialized, rate-mismatched subsystems}. The core challenge is capturing the system-level coordination and backpressure required to balance these two stages under dual SLOs.

\sys is designed to model these intricate system dynamics. For any given PD configuration (i.e., a fixed number of PD instances with specific parallelism), \sys simulates the end-to-end request lifecycle with high fidelity through a stateful, event-driven workflow:
(1). \sys models the prefill stage as a producer. When requests arrive, the \texttt{GlobalController} routes them to the prefill stage.
The \texttt{ClusterScheduler} and \texttt{ReplicaWorker} simulate its queuing and execution. Upon completion, the worker signals to the \texttt{Global} \texttt{Controller}, which transitions the request's state to \nolinkurl{PREFILL\_COMPLETE}. At this point, the generated KV-Cache is conceptually held in the prefill stage's memory buffer. 
(2). The decode stage is modeled as a consumer with a finite resource: GPU memory for KV-Caches. The \texttt{ClusterScheduler} of the decode stage continuously tracks its memory utilization. When a decoding request completes and its KV-Cache is evicted, the scheduler signals its updated memory availability to the \texttt{GlobalController}.
(3). The \texttt{GlobalController} acts as the central coordinator that respects backpressure. It maintains a queue of \nolinkurl{PREFILL\_COMPLETE} requests. It will initiate a \nolinkurl{KV_CACHE_TRANSFER} event for a request.

\noindent\textbf{AF disaggregation workflow simulation.}
The core simulation challenge, as highlighted by systems like MegaScale-Infer and Step-3, is to \textit{accurately capture the critical path of a multi-stage, micro-batch-driven workflow}, where even small imbalances between stages can create significant performance-degrading pipeline bubbles.

\sys addresses this by simulating the AF workflow as an event dependency graph. For any given AF configuration, \sys models the generation of a single token by orchestrating a complex graph of fine-grained events.
(1). The simulation begins when the \texttt{GlobalController} initiates a decode step. The \texttt{ReplicaWorker} in the decode-attn stage first partitions this global batch into a series of m simulated micro-batches.
(2). \sys's \texttt{GlobalController} and \texttt{ClusterScheduler}s dynamically construct a dependency graph for all operations across L model layers. 
\sys's event-driven engine processes this graph by scheduling events as soon as their dependencies are met. This inherently simulates the overlap: while \nolinkurl{A_TO_F_TRANSFER(i, k)} is in flight, the simulator can schedule \nolinkurl{ATTN_COMPUTE(i+1, k)} on the now-free attention GPU, perfectly capturing the latency-hiding principle of the ping-pong pipeline.
(3). The total time to generate one token is determined by the timestamp of the final event in the graph—typically \nolinkurl{FFN_COMPUTE(m, L)}—completing. 

\vspace{0.1em}
\noindent\textbf{MoE inference workflow simulation.}
Simulating MoE inference introduces a unique challenge: \textit{modeling a dynamic, data-dependent workflow} where performance is dictated not by average-case behavior, but by the worst-case straggler caused by token load imbalance.

\sys addresses these challenges by decomposing the MoE layer execution into a detailed, multi-step micro-workflow within the \texttt{ReplicaWorker}, ensuring that the effects of imbalance are modeled with high fidelity.
(1). \sys first configures the virtual model sharding to satisfy the system's topological constraints (e.g., \nolinkurl{attn_dp * attn_tp == moe_tp * moe_ep}).
(2). When the \texttt{ExecutionPredictor} encounters an MoE layer, it simulates the following sequence of events, explicitly tracking the consequences of token routing. The simulation first models the GEMM for the gating network. Subsequently, a pluggable routing module is invoked. \sys simulates the routing decision to generate a token-to-expert assignment map for the current batch. With the assignment map, the simulation of expert computation becomes heterogeneous. 
The \texttt{ExecutionPredictor} queries GroupedGEMM performance model with the actual number of tokens assigned to it for each expert i.
(3). After that, \sys simulates synchronization and straggler effects. \sys models the implicit synchronization barrier by calculating the latency as \nolinkurl{max[T_expert1, T_expert2, ..., T_expertN]}.




%% file: evaluation.tex

\begin{figure}[t] 
    \centering 
    \includegraphics[width=0.96\linewidth]{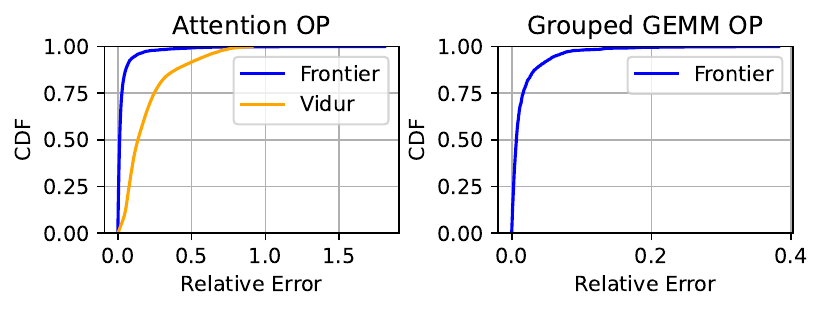} 
    \vspace{-2mm}
    \caption{
    CDF of the relative error in simulated operator runtime under dynamic workloads.}
    \label{fig:op_cdf}
\end{figure}

\begin{table}
    \renewcommand{\arraystretch}{0.9}
    \centering
    \footnotesize
    \begin{tabular}{ccccc}
        \toprule
        \raisebox{1.5ex}[0pt][0pt]{\textbf{Batch Size}} & 
        \raisebox{1.5ex}[0pt][0pt]{\textbf{Avg Input}} & 
        \raisebox{1.5ex}[0pt][0pt]{\textbf{Output}} & 
        \textbf{\shortstack[c]{Profiled \\ throughput}} & 
        \textbf{\shortstack[c]{Predicted \\ throughput}} \\
        \midrule
        4   & 32  & 1024  & 111.355 & 90.498 \\
        8   & 128  & 256  & 131.831 & 109.366 \\
        16   & 256 & 128 & 151.425 & 127.157 \\
        32 & 32 & 128 & 313.236 & 240.743 \\
        \bottomrule
    \end{tabular}
    \caption{End-to-end performance (throughput in tokens/s/GPU).}
    \label{tab:e2e_performance}
\end{table}

\section{Preliminary Evaluation} 
\label{sec:evaluation} 

\noindent\textbf{Setup.} Experiments are performed on an 8-GPU node featuring NVIDIA A800-SXM4-80GB GPUs with 400 GB/s NVLink interconnects. 
The profiling and training environment is configured with PyTorch 2.3, CUDA 12.1, Ray 2.42.1, and FlashInfer 0.1.6. 
End-to-end evaluation is conducted using vLLM 0.10.1 with the SharedStorageConnector KV interface. We use the Qwen2-7B-Instruct model~\cite{qwen2}.

\noindent\textbf{Operator accuracy.}
We evaluate the accuracy of Frontier on two critical operators: \at and \Gg. These operators are highly sensitive to variable inputs and prone to inaccuracies. For the \at operator, as shown in Figure~\ref{fig:op_cdf}, Frontier consistently outperforms Vidur, achieving significantly lower relative errors, with over 94\% of cases falling below 10\%.
For the \Gg operator, which is not supported by Vidur, we report results solely for Frontier. Frontier demonstrates high accuracy, with over 95\% of errors remaining below 6\%. 


\noindent\textbf{End-to-End accuracy.} We validate the end-to-end accuracy of \sys by simulating a PD disaggregated system with a 1:1 ratio of prefill to decode instances. As shown in Table~\ref{tab:e2e_performance}, we compare the predicted system output throughput against the profiled performance of a real system across a range of batch sizes and sequence lengths. The results demonstrate that \sys captures performance trends, with the predicted throughput (in tokens/s/GPU) consistently falling within a 19.0\% to 23.2\% relative error margin across tested cases.

%% file: conclusion.tex
\vspace{-0.2em}
\section{Discussion and Conclusion}
We introduced \sys, a high-fidelity simulator tailored for emerging LLM inference architectures, including disaggregated and MoE systems. Future work will expand on modeling core operators, quantifying simulation fidelity and cost, and demonstrating \sys's utility through diverse case studies for large-scale system design and optimization.